\documentclass{article}
\usepackage{nips15submit_e,times}
\usepackage{caption}
\usepackage{booktabs}
\usepackage{subcaption}
\usepackage{url,latexsym,enumitem}
\usepackage{amsmath,amsthm,amsfonts,mathtools}
\usepackage{graphicx,tabularx,bm,changepage,placeins}
\usepackage[normalem]{ulem}
\usepackage{color,colortbl}

\newcommand{\remove}[1]{}

\newcommand{\T}{\top}
\newcommand{\Lim}[1]{\raisebox{0.5ex}{\scalebox{0.8}{$\displaystyle \lim_{#1}\;$}}}
\definecolor{lightgray}{gray}{0.96}
\definecolor{darkgray}{gray}{0.7}
\definecolor{darkergray}{gray}{0.0}

\newcommand{\m}[1]{\multicolumn{2}{c|}{#1}}
\newcommand{\mno}[1]{\multicolumn{2}{c}{#1}}
\newcommand{\mm}[1]{\multicolumn{4}{c}{#1}}
\newcommand{\mmL}[1]{\multicolumn{4}{c|}{#1}}
\newcommand{\widebar}{\overline}
\renewcommand{\th}[0]{$^{\text{th}}$}
\renewcommand{\eqref}[1]{ (\ref{#1})}
\newcommand{\newcite}{\cite}
\usepackage{dcolumn}
\newcolumntype{H}{>{\setbox0=\hbox\bgroup}c<{\egroup}@{}}

\title{Sublinear Partition Estimation}
\author{Pushpendre Rastogi \qquad Benjamin Van Durme \\
  Johns Hopkins University}
\date{}
\nipsfinalcopy
\begin{document}
\maketitle
\begin{abstract}
  The output scores of a neural network classifier are converted to
  probabilities via normalizing over the scores of all competing
  categories.  Computing this \emph{partition function}, $Z$, is then
  linear in the number of categories, which is problematic as
  real-world problem sets continue to grow in categorical types, such
  as in visual object recognition or discriminative language modeling.
  We propose three approaches for sublinear estimation of the partition
  function, based on approximate nearest neighbor search and
  kernel feature maps and compare the performance of the proposed approaches
empirically.
\end{abstract}

\section{Introduction}
Neural Networks (and log-linear models) have out-performed other
machine learning frameworks on a number of difficult multi-class classification tasks
such as object recognition in images, large vocabulary speech
recognition and many others~\cite{russakovsky2014imagenet,hinton2012deep,zhang2000neural}.
These classification tasks become ``large scale'' as the number of
potential classes/categories increases. For instance,
discriminative language models need to choose the most probable word
from the entire vocabulary which can have more than 100,000 words in case of the
English language. And in the field of computer vision, the latest
datasets for  object recognition contain more than 10,000 object
classes and the number of categories is increasing every year~\cite{deng2010does}.

In certain applications neural networks may be used as
sub-systems of a larger model in which case it may be necessary to
convert the unnormalized score assigned to a class by a neural
network to a probability value.
To perform this conversion we need to compute the so-called
\textit{Partition Function} of a neural network which is just a
sum of the scores assigned by the neural network to all the
classes. Let $N$ be the number of output classes and
let  $q, v_i \in \mathbb{R}^d$ where $v_i$ represents weights of $i$th class.
Also, let $u_i$ be the score of the $i$\th{} class, i.e. $u_i = v_i \cdot q$
then the partition function $Z(q)$ is defined as:
\begin{align}
  Z(q) &= \sum_{i=1}^N\exp(u_i)
\end{align}
The problem of assigning a probability to the most probable
class can be stated as:
\begin{align}
  \text{Find, }\hat{\imath} &= \arg\max_i u_i \label{eq:mips}\\
  p(\hat{\imath}) &= \frac{\exp(u_{\hat{\imath}})}{Z(q)} \label{eq:prob}
\end{align}
Recently,
\cite{shrivastava2014alsh,neyshabur2014alsh,bachrach2014speeding}
have presented methods for solving~\eqref{eq:mips} by building upon
frameworks for performing fast randomized nearest
neighbor searches such as Locality Sensitive Hashing and Randomized k-d
trees. This paper presents methods for estimating the
value of $Z(q)$ required in~\eqref{eq:prob}.
\remove{We  note that the form of partition function that we are
interested in is specific to
artificial neural networks used in machine learning and has more
structure than the general
form used in statistical mechanics, $Z = \sum_s
\exp(-E(s)/kT)$. For us $E(s)$ is a linear
projection of the input vector $q$. While~\cite{Baxter1982exactly} presents a
number of algorithms for estimating the partition function that have
used for calculating partition functions for physical systems. None of
them apply to our setting because the dimensionality, $d$, of our data
is usually much greater than 3.}
While brute force parallelization is one effective strategy for
reducing the time needed for this computation our goal is to
estimate the partition function in asymptotically lesser runtime.
\section{Previous Work}
\label{sec:prev}
A number of techniques have been used previously for speeding
up the computation of the partition function in artificial neural networks. Computational efficiency of the normalizing step seems to be especially
important for these tasks because of
the large size of vocabulary needed to achieve state of the art
performance in these tasks. The prior work can be categorized by the
technique it uses as follows:
\par\textbf{Importance Sampling}:
The partition function can be written as $Z = N \mathbb{E}[\exp(u_i)]$
where the distribution of $i$ is uniform
  over $[1,\ldots,N]$. However, the attempts to estimate $Z$ by simply
  drawing $k$
  samples from the uniform distribution and replacing the expectation
  by its sample estimate are marred by the high variance of the
  estimate. \newcite{bengio2008adaptive} were the first
  to use
  importance sampling for reducing the variance of the sample average
  estimator. Their aim was to speed up the training of neural language
  models and they used an $n$-gram language model as the proposal distribution.
  Though they do not use their method for actually computing the
  partition function at inference time, their method
  could be easily extended for that purpose. The problem however with
  their method is that it
  requires the use of an external model for constructing the proposal
  distribution. Such an external model requires extra engineering and
  knowledge, specific to the problem domain, that may not be available.
\par\textbf{Hierarchical decomposition}:~\cite{mnih2008scalable}
introduced a method for breaking the original $N$-way decision problem
into a hierarchical one such that it
requires only $O(\log(N))$ computations. That method requires a change in the model
from a single decision problem into a chain of decision problems
computed over a tree. Also, since there is no a priori single most preferable way of
growing a tree for doing this computation therefore an external model is needed for
creating the hierarchical tree.
\par\textbf{Self-Normalization}: Some of the prior work side-steps the problem
of computing the partition function and trains neural networks with the added
constraint that the partition function should remain close to $1$ for inputs
seen during test time.

\newcite{MniTeh2012a} used Noise Contrastive Estimation\footnote{Please see section~\ref{sec:back} for a brief overview and section~\ref{ssec:mince} for a detailed explanation of NCE.}
with a heuristic to clamp the values of $Z$ to $1$ during training.
They demonstrated empirically that by doing so
the partition function at test time also remained close to $1$
though they did not provide
any theoretical analysis on how close to $1$ the value of the partition function
remains at test time.

On the other hand, \newcite{devlin2014fast} added a penalty term $\log{(Z(q))}^2$ to the
training objective of the neural network and empirically demonstrated on a large scale
machine translation task that they do not suffer from a large loss in accuracy even if they assume
that the value of the partition function is close to $1$ for all inputs at test time.
Recently, \newcite{andreas2015when} showed that after training on $n$ training examples,
with probability $1 - \delta$ the expected value of $\log(Z(q))$
lies in an interval of size $4\left(\sqrt{\frac{dN\log(dBRn) + \log(\frac{1}{\delta})}{2n}} + \frac{1}{n} \right)$
centered at $\frac{\sum_{i=1}^n \log(Z(q_i))}{n}$.
Here $N$ and $d$ are the number of output classes and number of
features in a loglinear model and $B$ and $R$ are upper bounds on
the infinity norm of $v_i$ and $q$ respectively.

\section{Background}
\label{sec:back}
\textbf{Maximum Inner Product Search (MIPS)}: MIPS refers to the
problem of finding points that have the highest inner product with an
input query vector.\footnote{Note that simply by
querying for $-q$ one can also find the vectors with the smallest inner
product.} Let us define $S_k(q)$ to be the set of
$k$ vectors that have the highest inner product with the vector $q$
and let us assume for simplicity that MIPS algorithms allow us to
retrieve $S_k(q)$ for arbitrary $k$ and $q$ in sublinear
time. The exact order of runtime depends on the
dataset and the indexing algorithm chosen for retrieval. For example, one could use the popular library FLANN
\cite{muja2009fast,muja2014scalable} or PCA-Trees\cite{sproull1991refinements} or LSH
itself\cite{dong2008modeling} for retrieving $S_k(q)$. Also,
\cite{he2012difficulty} presented a measure of hardness
of the dataset for nearest neighbor algorithms.

\cite{shrivastava2014alsh,anshumali2014asymmetric} and
\cite{neyshabur2014alsh} presented
methods for MIPS based on Asymmetric Locality Sensitive Hashing (LSH)
i.e., they used two separate hash functions
for the query and data. A different approach that
relies on reducing the problem of performing maximum inner product
search over a set of $d$-dimensional vectors to the problem of
performing nearest neighbor search in Euclidean distance over a set of
$d+1$ dimensional vectors was presented by
\cite{bachrach2014speeding,neyshabur2014alsh}. These algorithms
enable us to draw high probability samples from the unnormalized
distribution over $i \in [1,\ldots,N]$ induced by $q$ and we will
use this property heavily for creating our estimators.

\par \textbf{NCE}: NCE was introduced by
\cite{gutmann2010noise} as an objective function that can be
computed and optimized more efficiently than the likelihood objective
in cases where normalizing a distribution is an expensive operation. In
the same paper they proved that the NCE objective has a unique maxima,
and that it achieves that maxima for the same parameter values that maximize
the true likelihood function. Moreover the normalization constant
itself can be estimated as an outcome of the optimization.
The NCE objective relies on at least a single sample being available
from the true distribution which may be unnormalized and a noise
distribution which should be normalized.

\par \textbf{Kernel Feature Maps}:
The function $\exp(v_i \cdot q)$ is  a kernel that
depends on the dot product of $v_i$ and $q$, therefore, $\exp(v_i \cdot
q)$ is a dot product kernel\cite{scholkopf2002learning,kar2012random}. Every kernel that
satisfies certain conditions\footnote{It is sufficient for a kernel to be
  analytic, and have positive coefficients in its Taylor expansion
  around zero. These conditions are satisfied by the $\exp$ function.}
also has an
associated feature map such that the kernel can be decomposed as
a countable sum of products of feature functions\cite{smola2001regularization}:
\begin{align*}
  \exists \lambda_j \phi_j : k(x,x') = \sum_{j\in \mathbb{N}} \lambda_j \widebar{\phi_j(x)} \phi_j(x')
\end{align*}
If the values of $\lambda_j$ decrease fast enough
then one could approximate the $\exp$ kernel up to some small
tolerance by a finite summation of its
feature maps as follows:
\begin{align*}
    \exp(v_i \cdot q) &\approx \sum_{j=1}^P \lambda_j \widebar{\phi_j(v_i)} \phi_j(q)
\end{align*}

\par \textbf{Log Normal Distribution of $Z$}:
If we assume that $q \sim \mathcal{N}(\mu, \Sigma)$ then $u_i$ is
also a normal random variable with distribution
$\mathcal{N}(v_i^\T \mu, v_i^\T\Sigma v_i)$ and $\exp(u_i)$ is
log-normal distributed. In this case, $Z(q)$ is the sum of $N$
dependent log-normal random variables. There is no analytical
formula known for the distribution of $Z$, however, in general it is
known that the distribution of $Z$ is governed by the distribution of
the $\max(u_i)$  when  $Z$ is high enough due to a result by
\cite{asmussen2008asymptotics}.\footnote{They show that $\Lim{x \to
    \inf}\frac{p(Z(q) > x)}{\bar{F}(x)}$ equals the number of $u_i$
  that have the highest variance and highest mean. Here $\bar{F}(x)$
  is the tail of a log-normal CDF.} This suggests that one
could reasonably estimate $Z$ when its value is high enough, by
exponentiating and then summing only the top few $u_i$.
Unfortunately, when the value of $Z$ is not very large, then all $u_i$ become
significant for calculating $Z$. E.g.\ consider the pathological case
that $|q| = 0$ which means that $u_i = 0\, \forall i \in [1, \ldots, N]$.
In practice, for most values of $q$, the value of $Z(q)$ is not large enough to ignore the
contributions due to the tail of $u_i$.

\section{Methods}
\subsection{MIMPS: MIPS Based Importance Sampling}
In Section~\ref{sec:prev} we discussed the importance sampling based
approach for estimating the partition function and pointed out that
 the work so far relies on the presence of an external model or
 proposal distribution that can produce samples from the
high probability region. However, by utilizing the algorithms for
solving MIPS problem, we can overcome problems of engineering proposal distributions,
since we can retrieve the set
$S_k(q)$ (See Section~\ref{sec:back}). A naive estimator, which we call Naive~MIMPS
or NMIMPS, that utilizes $S_k(q)$ is the following:
\begin{align}
   \hat{Z}_{NMIMPS} = \sum_{s \in S_k(q)} \exp(s \cdot q)
\end{align}
Unfortunately NMIMPS requires $k$ to be very high and is not
realistic. Let $U_l$ represent a set of $l$ vectors sampled uniformly from
amongst the vectors that are not in $S_k$ then a better way of estimating $Z$ is:
\begin{align}
   \hat{Z}_{MIMPS} &= \sum_{s \in S_k(q)} \exp(s \cdot q) + \frac{N-k}{l} \sum_{u \in  U_l} \exp(u \cdot q)
\end{align}
In effect we are assuming that the values at the tail end of the probability
 distribution lie in a small range and thus a small sample size still
 has a small variance. A better estimator could be created by
 modeling the tail of the probability distribution, perhaps as a power law curve.

 \subsection{MINCE: MIPS Based NCE}
\label{ssec:mince}
 NCE is a general parameter estimation technique that can be used any
 where in place of maximum likelihood estimation. Specifically if we
 consider the values of the partition function to be a  parameter of
 the unnormalized distribution over $i$ induced by $q$ then by
 generating samples from the true distribution and a noise
 distribution ideally we can estimate it.
Since NCE requires samples from the true distribution which we can
generate by querying for $S_k(q)$ therefore methods that perform MIPS can be used for
estimating the value of $Z(q)$ as well. If our noise distribution is
uniform over the $N-k$ vectors not present in $S_k$ then the NCE
objective is; $\hat{Z}_{\text{MINCE}} = \arg\max_Z J(Z)$, where:
\begin{align}
  \text{where }J(Z) &= \sum_{s \in S_k(q)}\log(\frac{\exp(s \cdot q)/Z}{\exp(s \cdot q)/Z +\frac{l}{k} \frac{1}{N-k}}) + \sum_{l \in U_l} \log(\frac{\frac{l}{k} \frac{1}{N-k}}{\exp(l \cdot q)/Z +\frac{l}{k} \frac{1}{N-k}})
\end{align}
It is worthy to note that if we let $a_s =(\exp(s \cdot q)k(N-k)/l$ and
analogously define $b_l$ then the objective simplifies into a very
convenient form shown in~\eqref{eq:ncesimple} of which even the
third derivatives can be found  efficiently. Efficient
computation of the third derivative utilized through Halley's method, leads
to considerable speedup during optimization compared to using only the
second derivatives and Newton's method.
\begin{align}
  -J(Z) &= \sum_{i=1}^T \log(Z/a_i +1) + \sum_{j=1}^N \log(b_j/Z +1) \label{eq:ncesimple}
\end{align}

We also briefly note that one way of estimating the partition function could be to assume a parameteric form on the output distribution and then to use Maximum likelihood estimation which is the most efficient estimator possible when the form of the distribution is known. However even though individual class scores $u_i$ follow the lognormal distribution it is not clear how one could use MLE for computing the partition function in our setting.
\subsection{FMBE: Feature Map Based Estimation}
In Section~\ref{sec:back} we sketched how kernels could be linearized
into a sum over products of feature maps. This decomposition of a
kernel can be utilized for speeding up the computation of $Z$ as
follows:
\begin{align}
  \sum_{i=1}^N \exp(v_i \cdot q) &\approx \sum_{i=1}^N \sum_{j=1}^P
  \lambda_j \widebar{\phi_j(v_i)} \phi_j(q) = \sum_{j=1}^P \phi_j(q)\left(\lambda_j \sum_{i=1}^N \widebar{\phi_j(v_i)}\right)\nonumber\\
  \text{Let: } \tilde{\lambda}_j &= \lambda_j \sum_{i=1}^N \phi_j(v_i)\; \text{then }\hat{Z}(q) = \sum_{j=1}^P \tilde{\lambda}_j \phi_j(q)
\end{align}
Essentially, one could precompute $\tilde{\lambda}_j$ during training
and reduce the $O(N)$ summation to $O(P)$ along with a constant
factor, say $p$, needed to compute $\phi_j$.
Note that even though computing $\tilde{\lambda}_j$ involves the mapping $\phi$ which in
general is unknown. Let us now detail how we would compute $\phi$.
Overall this scheme would lead to
savings in time if $Pp < Nd$. Although~\cite{smola2001regularization}
gave explicit formulas for deriving the
eigenvalues $\lambda_j$ and eigen-functions $\Phi_j$ in terms of
spherical harmonics, unfortunately, we are not aware of any method for
 efficiently computing the spherical harmonics in high
 dimensions. Instead we will rely on a technique developed by
~\cite{kar2012random} for creating a randomized
 kernel feature map for approximating the dot product kernel as follows:
 \begin{align}
   \text{Let, }\phi_j(x) &= \sqrt{a_{M}p^{M+1}}\prod_{r=1}^M \omega_r^\top x\\
   \text{Then, } \exp(x,y) &\approx \sum_{j=1}^P\phi_j{(x)}^T \phi_j(y)
 \end{align}
Here $p$ is a hyper-parameter, usually taken to be 2. $a_m =\frac{1}{m!} $ is the
$m$th coefficient in the taylor expansion of $\exp$ and $M$ is chosen
by drawing a sample from a geometric distribution
$p[M=m]=\frac{1}{p^{m+1}}$ and
$\omega_r$ is a binary random vector each coordinate of which is
chosen from $\{-1, 1\}$ with equal chance. Refer to
\cite{kar2012random} for details\footnote{\cite{kar2012random} also present one more algorithm for creating random feature maps that we would not discuss here.}.

\section{Experiments}
We want to answer the following questions: (1) As a function of $k$,
if we have access to a system that can retrieve $S_k$
\footnote{We defined $S_k(q)$ to be the set of
$k$ vectors that have the highest inner product with the vector $q$ in Section~\ref{sec:back}}
then what
accuracy can be achieved by the proposed algorithms? (2) For a given
$k$, how does the accuracy then change in the face of error in
retrieval (such as would result from the use of an approximate nearest
neighbor routine such as MIPS)? (3) What is the accuracy of our proposed methods
as opposed to existing methods for estimating the partition function?

\subsection{Oracle Experiments}
\begin{figure}[ht]
   \centering
   \includegraphics[width=0.8\linewidth]{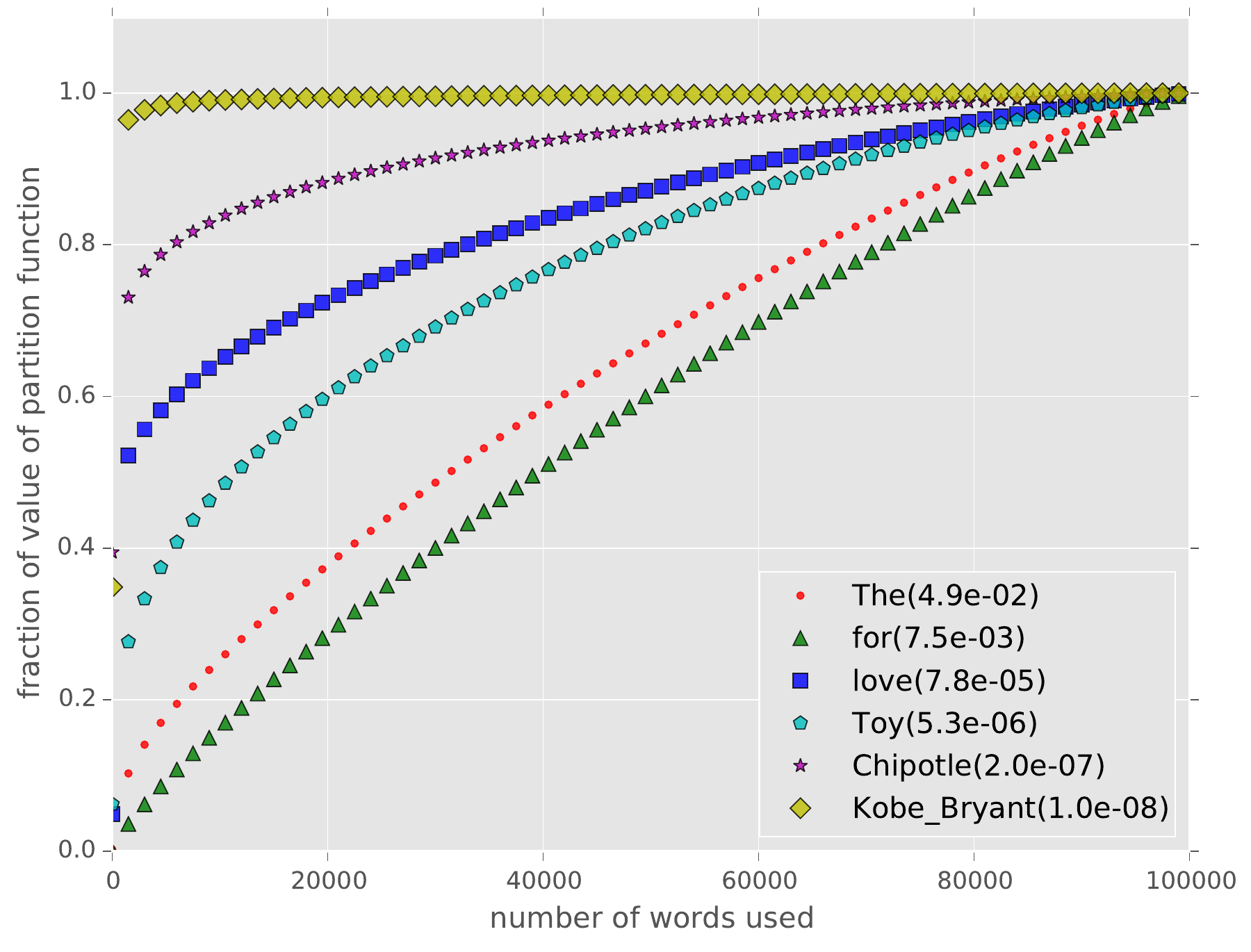}
   \caption{CDF over vocabulary items sorted in descending order from
     left to right according to their individual contribution to the
      distribution. Every curve is associated with a distinct context
      word marked in the legend. The bracketed numbers in the legend indicate the
     frequency of occurrence of these words in an English Wikipedia
     corpus. We can see that high frequency, common words tend to induce
     flat distributions.}
\label{fig:mikocdf}
 \end{figure}

In Section~\ref{sec:back} in the discussion of the log-normal distribution of $Z$
we explained how the number of neighbors
needed for estimating $Z$ was dependent on the value of $Z$ itself.\footnote{Also see figure~\ref{fig:mikocdf}}
Our first set of experiments relies on real-world, publicly available collection of
vectors: the neural word embeddings dataset released by
\cite{mikolov2013distributed} that consists of 3 million,
$300$ dimensional vectors, each representing a distinct word or phrase
trained on a, 100 billion token, monolingual corpus of news
text.\footnote{URL:\url{code.google.com/p/word2vec}} Each vector
represents a single word or phrase. More pertinently, the dot product
between the vectors $v_i, v_j$, associated to the vocabulary items
$w_i, w_j$ respectively, represents the unnormalized log probability
of observing $w_i$ given $w_j$:
 \begin{align}
   p(w_i|w_j) = \frac{\exp(v_i \cdot v_j)}{\sum_{k=1}^N \exp(v_k \cdot v_j)}
 \end{align}
For our experiments we used the first $100,000$ vectors from the
3 Million word vectors and all experiments in this subsection are on this
set.  Note that we do not normalize the vectors in any way: this
 ensures that we stay true to real-world situation in which vectors are
 the weights of a trained neural network, and consequently we can not
 modify them.

 In Figure~\ref{fig:mikocdf} we show CDFs over words given context,
 sorted such that the words contributing the highest probability
 appear to the left. We can see that less than $1000$ nearest
 neighbors (in terms of largest magnitude dot-product) are needed for
 recovering 80\% of the true value of the partition function for the
 rare words \textit{Chipotle} and \textit{Kobe\_Bryant}, but close to
 80K neighbors are needed for common words that have high frequency of
 occurrence in a monolingual corpus. This is explained by the fact
 that common terms such as ``The'' occur in a wide variety of contexts
 and therefore induce a somewhat flat probability distribution over
 words. These patterns indicate that the Naive MIMPS estimator would
 need an unreasonably large number of nearest neighbors for correctly
 estimating the partition function of common words and therefore we do
 not experiment with it further and focus on MIMPS, MINCE and FMBE\@.

We implemented MIMPS, MINCE and FMBE based on an oracle
ability to recover $S_k$, to which we then add errors in a
deterministic fashion.  The resultant estimates of $Z$ are then
tabulated based on their mean absolute relative
error.\footnote{Percentage Absolute Relative error $\mu$ = $100|\frac{\hat{Z}-Z}{Z}|$}.

Our query set consists of $10,000$ items taken from across the
top $100,000$ vectors chosen initially. Each query represents the context (features) that
are best ``classified'' by one of the many categories.  In the case of
word-embeddings and language models, this would be some preceding word
context which would be extracted and used in measuring the surprisal
of the next word in a sequence.\footnote{Surprisal being a function of
the probability assigned by the model to an observation given
context, and that probability assignment requiring the computation
of $Z$.}  We simulate this context by taking the representation of a
given item from the vocabulary (a query vector) and randomly adding
varied levels of noise with controlled relative norms. Every
experimental setting was ran three times with different seeds to
maintain a low standard error.

Table~\ref{tab:l} presents the hyper-parameter tuning results for the
different algorithms (UNIFORM, MIMPS and MINCE). We can see a symmetric behavior in the table for
MIMPS which is surprising and we can see that the uniform case (which we
model as a special case of MIMPS where k=0) performs badly. It is good to
see that at $k=1000$ and $l=1000$ the error in $Z$ is quite low but more
exciting to see that when $k=100$ and $l=100$ then the error is only 7.1\%
with only 0.1\% standard error. This means that by retrieving only
0.1\% of the original vocabulary one can reasonably estimate the value
of $Z$ with low error. The MINCE estimator and the FMBE estimators do
not fare well although the decrease in error of the MINCE algorithm as
the number of noise samples is increased agrees with intuition.
The FMBE algorithm had  $\mu=100$ at $D=10000$  and $\mu=83.8$ at
$D=50000$. The standard error in both cases was lower than
$0.1$. Clearly the FMBE algorithm would require far higher number of dimensions
in the feature map created through random projections before giving
reasonable results and it might be better to experiment with the newer
methods for generating kernel feature maps that come with better
theoretical guarantees, e.g.\ by~\cite{pham2013fast}. We defer this investigation to future work.

\begin{table}[htbp]
\centering
{
\begin{tabular}{l|rr|rr|rr}
 & \m{l=1000}   & \m{l=100} &   \mno{l=10}   \\
 & $\mu$ & $\sigma$ & $\mu$ & $\sigma$ & $\mu$ & $\sigma$\\\hline
Uniform        & 101.8   & 3.1    & 117.3   & 10.4  & 97.3   & 10.5 \\
MIMPS (k=1000) & 0.8     & 0.0    & 2.7     & 0.0   & 8.2    & 0.1  \\
MIMPS (k=100)  & 2.4     & 0.0    & 7.1     & 0.1   & 16.1   & 0.2  \\
MIMPS (k=10)   & 8.1     & 0.1    & 17.1    & 0.3   & 27.4   & 0.7  \\
MIMPS (k=1)    & 28.7    & 0.6    & 39.3    & 2.1   & 47.0   & 2.7  \\
MINCE (k=1000) & 96285.4 & 2124.1 & 12413.0 & 363.9 & 2527.3 & 72.9 \\
MINCE (k=100)  & 3780.9  & 125.8  & 667.4   & 20.2  & 846.5  & 5.1  \\
MINCE (k=10)   & 230.9   & 7.9    & 330.3   & 2.1   & 827.1  & 5.0  \\
MINCE (k=1)    & 133.7   & 0.8    & 317.3   & 2.0   & 525.2  & 3.5  \\
\end{tabular}
}
\caption{Mean absolute relative error, $\mu$, and their associated
  standard error, $\sigma$, for different algorithms at
  varying settings of the hyper-parameters $k$ and $l$ that govern the
  number of vectors retrieved.}
\label{tab:l}
\end{table}

Table~\ref{tab:nt} shows the results of adding noise to the query vectors.
As we mentioned before and interesting experiment for us was to
restrictively simulate the type of errors that these
 estimators might encounter in a real setting where the vector with
 the highest or second highest inner product might not be made
 available to the estimators. We tabulated the performance of the
 estimators on these type of errors in Table~\ref{tab:rt}. It is
 disconcerting to see the huge increase in error when
 the most important neighbor is absent from the retrieved set and
 clearly the importance of neighbors decreases as their rank
 increases. This indicates that one should use retrieval mechanism
 that have a high chance of retrieving the single best nearest
 neighbor in practice. This evidence is important while deciding
 between different indexing schemes that solve the MIPS problem.

\begin{table}[htbp]
  \centering
{
\begin{tabular}{l|rr|rr|rr|rr}
        & \m{noise=0\%}  & \m{noise=10\%}  & \m{noise=20\%}  & \mno{noise=30\%}  \\
& $\mu$ & $\sigma$ & $\mu$ & $\sigma$ & $\mu$ & $\sigma$& $\mu$ & $\sigma$ \\\hline
Uniform & 101.8 & 3.1      & 103.6 & 3.1      & 104.1 & 3.1 & 105.0 & 3.1 \\
MIMPS   & 0.8   & 0.0      & 0.9   & 0.0      & 0.9   & 0.0 & 0.9   & 0.0 \\
MINCE   & 230.9 & 7.9      & 229.9 & 7.9      & 233.7 & 8.0 & 231.5 & 8.4 \\
FMBE    & 83.8  & 0.2      & 85.2  & 0.2      & 85.8  & 0.2 & 87.1  & 0.2 \\
\end{tabular}
}
  \caption{Results at varying levels of gaussian noise added to the
    query vectors to make them deviate from the actual. The header of
    the column indicates the norm of the noisy vector relative to the
    norm of the original vector. $K$ and $L$ were both set to $1000$
    for MIMPS and to $1$ and $1000$ for MINCE.}
\label{tab:nt}
\end{table}

\begin{table}[htbp]
  \centering
{
\begin{tabular}{l|rr|rr|rr|rr}
 & \m{ret err=None}   & \m{ret err=1}   & \m{ret err=2} &  \mno{ret err=[1 2]}  \\
& $\mu$ & $\sigma$ & $\mu$ & $\sigma$ & $\mu$ & $\sigma$& $\mu$ & $\sigma$ \\\hline
MIMPS & 0.8 & 0.0 & 39.3 & 0.2 & 6.1 & 0.0 & 45.0 & 0.2 \\
MINCE & 133.7 & 0.8 & 133.7 & 0.8 & 133.7 & 0.8 & 133.7 & 0.8 \\
\end{tabular}
}
  \caption{The performance of the estimators with simulated retrieval
    errors in the oracle system. ``ret err=None'' represents no error
    where ``ret'' stands for retrieval and ``ret err=1'' represents that the most closest vector in terms
  of innre product was missing from the $S_k$ retrieved by the oracle
  and ``ret err=0 1'' means that the first and second items were
  missing. We can see that the error increases as more and more items
  go missing.$K$ and $L$ were both set to $1000$ for MIMPS and to $1$ and $1000$ for MINCE.}
\label{tab:rt}
\end{table}

\subsection{Language Modeling Experiments}
We now move beyond controlled experiments and do an end-to-end experiment by training a log bilinear
language model~\cite{mnih2008scalable} on text data from sections 0--20 of the Penn Treebank Corpus.
At test time we estimate the value of the partition function
for the contexts in sections 21--22 of the Penn Treebank and compare the approximation
to the true values of the partition function.
We train the log-bilinear language models using NCE and
clamp the value of the partition function to be one
while training the language model which enables us to do evaluate
the accuracy of our method against the most common usage of NCE for language modeling.
For the following experiments we use the method MIMPS that we implement using the specific MIPS algorithm
presented by~\cite{bachrach2014speeding} that in turn is implemented by
modifying the implementation of K-Means Tree in FLANN~\cite{muja2009fast}.

Remember that our goal is to estimate the true value of the partition function in the test corpus.
There are two main hyper parameters in our approach, the number of ``head'' samples
and the number of ``tail'' samples.
We train the LBL language model with the dimensionality of 300 and context size of 9
and tabulate the results
as the number of head and tail samples is varied
in table~\ref{tab:flann}. We can see that with around 100
head samples and 100 tail samples the estimation accuracy becomes
better than the heuristic of assuming that the value of $Z$ is $1$.

\begin{table}[htbp]
{\addtolength{\tabcolsep}{-3pt}
  \begin{tabular}{l|cccc|cccc}
        & \mmL{$l=10$} & \mm{$l=100$} \\\hline
        & {AbsE-MIPS}  & {AbsE-NCE} & {\%Better}  & {Speedup} & {AbsE-MIPS} & {AbsE-NCE} & {\%Better} & {Speedup} \\\hline
$k=10$  & 1063.5 & 352 & 34   & 18.5 & 728.5 & 352 & 47.5 & 13.5 \\
$k=50$  & 989.5  & 352 & 46.5 & 16   & 554   & 352 & 61.5 & 13   \\
$k=100$ & 229    & 352 & 55.5 & 14.5 & 198.5 & 352 & 70.5 & 10   \\
  \end{tabular}}
  \caption{AbsE column contains the total absolute difference between the estimated value
of the partition function and the true value over the test set (Section 21--22 of the Penn Treebank Corpus)
 for the corresponding estimators. The test set contained close to $10,000$ contexts.
\%Better refers to the number of times the MIPS estimator gives a better estimate than the NCE heuristic as a percentage of the total number of contexts in the test set.
The Speedup refers to the speedup achieved over brute force
computation by the corresponding MIPS method. }
\label{tab:flann}
\end{table}

\section{Conclusions}
We presented three new methods to estimate the partition function of a neural
network or a log-linear model  using recent algorithms from the
field of randomized algorithms for nearest neighbor search, new statistical estimators and
randomized kernel feature maps. We found
that it is possible to compute the true value of the partition
function with a small number of samples both under ideal
conditions where we have an oracle for retrieving the true set k
vectors closest to a query vector and on at least one real dataset using
the algorithm for MIPS described in~\cite{bachrach2014speeding} implemented using the
FLANN toolkit. We also noted that
the estimator MIMPS seems to be the most reasonable way to do
so. Initially we were hopeful that  the MINCE estimator could also be
successfully used but we found that it did not work so well.

While the data used for our experiments was always a real world dataset,
we performed both controlled experiments where settings such as retrieval error and the
creation of query vectors were carefully controlled to tease apart the
sources of errors and end-to-end tasks.
Based on the control experiments we can see that the performance of the algorithms
critically depend on
the indexing mechanism employed and it might be possible to extend some of the
guarantees of those algorithms to our problem by using the results
described in~\cite{he2012difficulty}.

We also note that while a theoretical analysis of the performance of an estimator of the
partition function would be extremely desirable, doing so for methods that
rely on LSH that would need a three step analysis:
(1) Analyze how the actual data (text or images) affects the weights
learnt on the outer layer of a neural network. This process is not
well understood.
(2) How that distribution of weights would affect the performance of
nearest neighbor retrieval. Perhaps the approach taken in~\cite{he2012difficulty}
could be extended for this purpose.
(3). Finally, how the error in Nearest neighbor retrieval would affect
the accuracy of the estimator. This analysis could be done by assuming some parameteric distribution
on the distributions of scores assigned to the output classes.
We defer solutions to one or more of these steps to future work.
\remove{\section*{Acknowledgments}
This material is based on research sponsored by Defense Advanced
Research Projects Agency (DARPA) under the Deep Exploration and
Filtering of Text (DEFT) Program (Agreement number FA8750-13-2-0017).}
\bibliographystyle{abbrv}
\bibliography{supar_nips}
\end{document}